\documentclass{article}

\usepackage{arxiv}

\usepackage[utf8]{inputenc} 
\usepackage[T1]{fontenc}    
\usepackage{hyperref}       
\usepackage{url}            
\usepackage{booktabs}       
\usepackage{amsfonts}       
\usepackage{nicefrac}       
\usepackage{microtype}      
\usepackage{lipsum}
\usepackage{graphicx}
\usepackage{float}
\usepackage{amsmath}
\usepackage{amsthm}

\usepackage{romannum}
\usepackage[en-US,showdow]{datetime2}
\usepackage{babel}
\graphicspath{ {./images/} }
\usepackage{tikz}
\usepackage{pgfplots}
\usepackage{pgfplotstable}
\usepackage[ruled,linesnumbered]{algorithm2e}
\usepackage{enumitem}
\usetikzlibrary{intersections,matrix,chains,positioning,decorations.pathreplacing,arrows,calc,intersections, datavisualization}

\title{Learning in Feedforward Neural Networks Accelerated by Transfer Entropy}

\author{
 Adrian Moldovan \\
 Department of Electronics and Computers\\
 Transilvania University\\
 \texttt{adrian.moldovan@gmail.com}\\
    \And
 Angel Ca\c taron \\
 Department of Electronics and Computers\\
 Transilvania University\\
 \texttt{cataron@unitbv.ro}\\
    \And
  R\u azvan Andonie\\
 Department of Computer Science\\
 Central Washington University\\
 \texttt{razvan.andonie@cwu.edu}\\
}

\begin{document}
\maketitle
\begin{abstract}
Current neural networks architectures are many times harder to train because of the increasing size and complexity of the used datasets. Our objective is to design more efficient training algorithms utilizing causal relationships inferred from neural networks. The transfer entropy (TE) was initially introduced as an information transfer measure used to quantify the statistical coherence between events (time series). Later, it was related to causality, even if they are not the same. There are only few papers reporting applications of causality or TE in neural networks. Our contribution is an information-theoretical method for analyzing information transfer between the nodes of feedforward neural networks. The information transfer is measured by the TE of feedback neural connections. Intuitively, TE measures the relevance of a connection in the network and the feedback amplifies this connection. We introduce a backpropagation type training algorithm that uses TE feedback connections to improve its performance.
\end{abstract}


\section{Introduction and Related Work}

We generally differentiate between statistical correlation and causality. Often, when correlation is observed, causality is wrongly inferred and we are tempted to identify causality through correlation. This is because of the inability to detect a time lag between a cause and effect, which is a prerequisite for causality \cite{Marwala2015}.

Following Shadish et al. \cite{Shadish2001}, the three key criteria for inferring a cause and effect relationship are (1) the cause preceded the effect, (2) the cause was related to the effect, and (3) we can find no plausible alternative explanation for the effect other than the cause.

According to \cite{Zaremba2014}, there is an important distinction between the ``intervention-based causality'' and ``statistical causality''. The first concept, introduced by Pearl \cite{Pearl2009}, combines statistical and non-statistical data and allows one to answer questions, like ``if we give a drug to a patient, i.e., intervene, will their chances of survival increase?'' Statistical causality does not answer such questions, because it does not operate on the concept of intervention and only involves tools of data analysis. The causality in a statistical sense is a type of dependence, where we infer direction as a result of the knowledge of temporal structure and the notion that the cause has to precede the effect. We will focus here only on statistical causality measured by the information transfer approach.

A relatively recent information transfer measure is the transfer entropy (TE). The TE was introduced by Schreiber \cite{Schreiber2000} not as a causality indicator, but as an information transfer measure used to quantify the statistical coherence between events (time series). For a comprehensive discussion of the TE vs. causality paradigms we refer to the work in \cite{Lizier2010}. In our previous work, we introduced the Transfer Information Energy (TIE) \cite{Cataron2017, Cataron2019} as an alternative to the TE. Whereas the TE can be used as a measure of the reduction in uncertainty about one event given another, the TIE measures the increase in certainty about one event given another.

Causality and information transfer are not exactly the same. Causality is typically related to whether interventions on a source have an effect on the target. Information transfer measures how observations of the source can predict transitions of the target. Causal information flow describes the causal structure of a system, whereas information transfer can then be used to describe the emergent computation on that causal structure \cite{Lizier2010}.

The directivity of information flow through a channel was defined by Massey \cite{Massey1990} in the form of directed information. The author shows that in the presence of feedback, this is a more useful quantity than the traditional mutual information. From a similar perspective, the TE measures the information flow from one process to another by quantifying the deviation from the generalized Markov property as a Kullback--Leibler distance, thus both TE and the directed information can be used to estimate the directional informational interaction between two random variables.

TE is a directional, dynamic measure of predictive information, rather than a measure of the causal information flow from a source and to a destination. To be interpreted as information transfer, the TE should only be applied to causal information sources for the given destination \cite{Lizier2010}. We will use the information transfer measured by the TE to establish the presence of and quantify causal relationships between the nodes (neurons) of neural networks.


In the current deep learning era, neural architectures are many times hard to train because of the increasing size and complexity of the used datasets. Our main question is how causal relationships can be inferred from neural networks. Using such relationships, can we define better training algorithms? There are very few results reporting applications of causality or transfer information in neural networks. We will refer to them in the following.


TE has been used for the quantification of effective connectivity between neurons \cite{Lizier2011, Vicente2011, Shimono2015, HuiFang2018}. To the extent of our knowledge, the work in \cite{Obst2010, Herzog2017} represent the only attempts to use TE for improving the learning capability of neural networks.

The reservoir adaptation method in \cite{Obst2010} optimizes the TE at each individual unit, dependent on properties of the information transfer between input and output of the system. It improves the performance of online echo state learning and recursive least squares online learning.

Causal relationships within a neural network are defined by F\'{e}raud et al. in \cite{Feraud2002}. To explain the classification obtained by a multilayer perceptron, F\'{e}raud et al. introduced the concept of ``causal importance'' and defined a saliency measurement allowing the selection of relevant variables. Combining the saliency and the causal importance allowed them an interpretation of the trained neural network.

Herzog et al. \cite{Herzog2017} used feedforward TE between neurons to structure neural feedback connectivity. These feedback connections are then used to improve the training algorithm in a convolutional neural network (CNN) \cite{Patterson2017}. In deep learning, a CNN is a class of deep neural networks, most commonly applied to image analysis. Intuitively, a CNN is a multilayered neural network that uses convolution in place of general matrix multiplication in at least one of its layers. Herzog et al. averaged (by layer and class of the training sample) the calculation of TE gathered from directly or indirectly connected neurons, using thresholded activation values to obtain the required time series. The averaged TEs are indirectly implied in the subsequent neuron's activations as part of the training with feedback. They are potentiated with a layer distance amplifier and the new value is summed to the input of the activation function. As a result, only one TE derived value is used for each of the layers. Herzog et al. made two interesting observations about why using TE for defining TE feedback in CNN networks:
  \begin{itemize}[leftmargin=*,labelsep=5.5mm]
    \item There is a decreasing feedforward convergence towards higher layers.
    \item The TE is in general lower between nodes with larger layer distances than between neighbors. This is caused by the fact that long-range TE is calculated by conditioning on the intermediate layers. Thus, there is a higher probability to form long-range as compared to short-range feedback connections.
  \end{itemize}


Our contribution is a novel information-theoretical approach for analyzing information transfer (measured by TE) between the nodes of neural networks. We use the information transfer to establish the presence of relationships and the quantification of these between neurons. Intuitively, TE measures the relevance of a connection in the network and the feedback amplifies this connection. We introduce a backpropagation-type training algorithm which uses TE feedback connections to improve its performance.

The paper has the following structure. In Section \ref{TE}, we introduce the formal definition of the TE and enumerate some of its applications. Section \ref{background} describes how the feedback TE can be numerically approximated during the training of a feedforward neural network. In Section \ref{our_approach}, we present our approach for integrating the TE as a feedback in the training algorithm of a neural classifier. The closest related work is Herzog's et al. paper \cite{Herzog2017}, and we will explain the differences between the two approaches. Section \ref{results} presents several experiments performed on a toy example and on standard benchmarks. Section \ref{discussion} analyzes the results of the numerical experiments. Section \ref{conclusions} contains final remarks. The Appendix \ref{trainings} presents further details of our experiments.

\section{Background: Transfer Information Entropy} \label{TE}
We start by introducing the formal definition of TE. A detailed presentation can be found in \cite{Bossomaier2016}. The connection between TE and causality in time series analysis is discussed in \cite{Hlavackova-Schindler2007}.

TE measures the directionality of a variable with respect to time based on the probability density function. For two discrete stationary processes $I$ and $J$, TE relates $k$ previous samples of process $I$ and $l$ previous samples of process $J$ and is defined as follows \cite{Schreiber2000, Kaiser2002},

\begin{equation}\label{eq:TEcond}
	TE_{J\rightarrow I}=\sum_{t=1}^{n-1}{p(i_{t+1},i_{t}^{(k)},j_{t}^{(l)}) \: log \frac{p(i_{t+1}|i_{t}^{(k)},j_{t}^{(l)})}{p(i_{t+1}|i_{t}^{(k)})}},
\end{equation}
where $i_t$ and $j_t$ are the discrete states at time $t$ of $I$ and $J$, respectively, and $i_t^{(k)}$ and {$j_t^{(l)}$} are the $k$ and $l$ dimensional delay vectors of time series $I$ and $J$, respectively. {The three symbols $i_{t+1},i_{t}^{(k)},j_{t}^{(l)}$ for computing probabilities are sequences of time series symbols.}

$TE_{J\rightarrow I}$ measures the extend to which time series $J$ influences time series $I$. The TE is asymmetric under the exchange of $i_t$ and $j_t$, and provides information regarding the direction of interaction between the two time series. With respect to mutual information, the TE can be interpreted as being equivalent to the conditional mutual information \cite{Hlavackova-Schindler2007}.

The accurate estimation of entropy-based measures is generally difficult. There is no consensus on an optimal way for estimating TE from a dataset \cite{Gencaga2015}. Schreiber proposed the TE using correlation integrals \cite{Schreiber2000}. The most common TE estimation approach is histogram estimation with fixed partitioning. This simple method is not scalable for more than three scalars. Moreover, it is sensitive to the size of bins used. Other nonparametric entropy estimation methods have been also used for computing the TE \cite{Hlavackova-Schindler2009,Gencaga2015,Zhu2015}: kernel density estimation methods, nearest-neighbor, Parzen, neural networks, etc.

The best known applications of TE are in financial time series analysis. TE was used to compute the information flow between stock markets or to determine relationships between stocks, indexes, or markets \cite{Kwon2008,Sandoval2014}. Other application areas are neuroscience, bioinformatics, artificial life, and climate science \cite{Bossomaier2016}.

\section{How to Compute the Feedback Transfer Entropy in a Neural Network} \label{background}
An interesting question is how to compute the TE in a neural network, as the TE was originally defined for time series. We describe in the following how the TE can be defined and computed in a feedforward neural network as a feedback measure.

A feedforward neural network is the simplest artificial neural network architecture. The information moves in only one direction, forward, from the input nodes, through the hidden nodes (if any) and to the output nodes. There are no cycles in the network. The most common training method of such a network is to use the output result backwards, to adjust the weights of the connections between nodes. In our case, we will adjust the weights also considering the TE feedback measure.

We denote by \emph{FF+FB} the network with a TE feedback, whereas a network without feedback is named \emph{FF} (feedforward only). $R$ is the number of epochs used in the training algorithm for both the \emph{FF+FB} and \emph{FF} networks, whereas $r$ is the index of a current epoch, $r \in \{1,\dots,R\}$.

The information transfer between two neurons can be computed if the outputs of the neurons are logged along the training process. After $p$ consecutive training steps in \emph{FF+FB}, we obtain two time series, of length $p$ each, with the observations aligned in time by the index of the training step and sample position. For the \emph{FF+FB} network trained with the backpropagation algorithm, we denote by $o_{i}^{r,n}$ the output of neuron $i$ in layer $l-1$ when processing the $n^{th}$ training sample during epoch $r$. Similarly, $o_{j}^{r,n}$ is the output of neuron $j$ in layer $l$. The layer index is not required here since we compute the information transfer only between pairs of neurons from adjacent layers.

The time series are obtained by discretization of the continuous values $o_{i}^{r,n}$ and $o_{j}^{r,n}$ (we use the sigmoid activation function). The continuous sub-intervals are mapped to discrete values using binning: $s_{i}^{r,n} = 1$ for $o_{i}^{r,n} > g$ and $s_{i}^{r,n} = 0$ for $o_{i}^{r,n} \leq g$, where $s$ stands for time \emph{series}. We record the time series only after the first 10 training patterns were processed. This value was obtained experimentally, optimized for smaller training sets.

We estimate the TE for the two generated time series by approximating the probabilities with relative frequencies. A higher number of discrete levels gives a better approximation of the TE, but also requires longer time series. Obviously, a reduced number of discrete values is computationally more efficient. We illustrate in Figure \ref{fig:tscompute} the computational pipeline that (1) collects process values from the data flow that goes through two connected neurons and (2) computes the TE between the two discrete time series produced from the output of each neuron. In the following, we denote by $te$ the computed (approximated) TE value.

\def\layersep{2.5cm}

\tikzset{%
  every neuron/.style={
    circle,
    draw,
  },
  neuron missing/.style={
    draw=none,
    execute at begin node=\color{black}$\vdots$
  },
}

\begin{figure}[H]
  \centering
  \begin{tikzpicture}[shorten >=1pt,->,draw=black, node distance=\layersep]

      \begin{scope}[scale=0.6][local bounding box=scope1]
        \tikzstyle{neuron}=[circle,draw,fill=none,minimum size=12pt,inner sep=0pt]

        \foreach \m/\l [count=\y] in {1,2,3,missing,4}
          \node [every neuron/.try, neuron \m/.try] (input-\m) at (0,2.5-\y) {};

        \foreach \m [count=\y] in {1,missing,2}
          \node [every neuron/.try, neuron \m/.try ] (hidden-\m) at (2,2-\y*1.25) {};

        \foreach \m [count=\y] in {1,missing,2}
          \node [every neuron/.try, neuron \m/.try ] (output-\m) at (4,1.5-\y) {};

        \foreach \l [count=\i] in {1,2,3,n}
          \draw [<-] (input-\i) -- ++(-1,0)
            node [above, midway] {};

        \foreach \l [count=\i] in {1,n}
          \node [above] at (hidden-\i.north) {};

        \foreach \l [count=\i] in {1,n}
          \draw [->] (output-\i) -- ++(1,0)
            node [above, midway] {};

        \foreach \i in {1,...,4}
          \foreach \j in {1,...,2}
            \draw [->] (input-\i) -- (hidden-\j);

        \foreach \i in {1,...,2}
          \foreach \j in {1,...,2}
            \draw [->] (hidden-\i) -- (output-\j);

        \node [neuron, thick] (x) at ($ (input-1) + (2.3, 1.9) $) {};
        \node [neuron, thick] (y) at ($ (hidden-1) +  (3, 1.5) $) {};


        \draw [thick, rotate=82, label={ME}] (output-1)
            circle [
              label={ME},
              x radius = 0.45cm,
              y radius = 1.7cm,
              xshift=0.8,
              yshift=30];

        \draw [->, thick] (x) -- (y);
        \draw [->, thick] ($(hidden-1) + (0.76, +0.36)$) to [bend left=20]($(x) + (1.3, -0.7) $);
        \draw [->, thick] (x) to [bend left=30]($(x) + (-1.3, 4.1) $);
        \draw [->, thick] (y) to [bend left=30]($(y) + (0.5, 1.0) $);

      \end{scope}

      \begin{scope}[local bounding box=comb1, shift={($(x.east)+(-1.0cm, 2.2cm)$)}]
        \node [color=red] (g) at (0,0.4cm) {$g$};
        \pgfplotstableread{
          x y
          0 0.591780304
          1 0.476430321
          2 0.499985927
          3 0.500072271
          4 0.496428962
          5 0.476190413
          6 0.615528519
          7 0.495910403
          8 0.49981736
          9 0.449868484
        }{\loadedtable}
        \begin{axis}[
            ycomb,
            height=5cm,
            width=5cm,
            unit vector ratio*={1 15 0.2},
            clip=false,
            axis lines=none,
            xtick=\empty, ytick=\empty, ztick=\empty
          ]

          \path [thick,-,color=red] (-0.3,0.53) edge (9.6, 0.53);


          \addplot+[mark size=1.5] table [
              x=x,
              y expr={\thisrow{y} >=0.53 ? \thisrow{y} : NaN},
          ] {\loadedtable};

          \addplot+[mark size=1.5] table [
              x=x,
              y expr={\thisrow{y} < 0.53 ? \thisrow{y} : NaN},
          ] {\loadedtable};

        \end{axis}
      \end{scope}

      \begin{scope}[local bounding box=comb12, shift={($(comb1.east)+(1.6cm, -0.40cm)$)}]
        \pgfplotstableread{
          x y
          0 1.
          1 0.
          2 0.
          3 0.
          4 0.
          5 0.
          6 1.
          7 0.
          8 0.
          9 0.
        }{\loadedtable}
        \begin{axis}[
            ycomb,
            height=5cm,
            width=4.5cm,
            unit vector ratio*={2 5 0.2},
            clip=false,
            axis lines=none,
            xtick=\empty, ytick=\empty, ztick=\empty
          ]

          \draw [ultra thin, -] (axis cs:\pgfkeysvalueof{/pgfplots/xmin},0.)
              -- node[left=1.6cm] {0} (axis cs:\pgfkeysvalueof{/pgfplots/xmax},0.);

          \draw [ultra thin, -] (axis cs:\pgfkeysvalueof{/pgfplots/xmin},1.)
              -- node[left=1.6cm] {1} (axis cs:\pgfkeysvalueof{/pgfplots/xmax},1.);


          \addplot+[mark size=1.5] table [
              x=x,
              y expr={\thisrow{y} >0. ? \thisrow{y} : NaN},
          ] {\loadedtable};

          \addplot+[mark size=1.5] table [
              x=x,
              y expr={\thisrow{y} <= 0. ? \thisrow{y} : NaN},
          ] {\loadedtable};

        \end{axis}
      \end{scope}

      \draw [->, thick] (comb1) to (comb12);

      \begin{scope}[local bounding box=comb2, shift={($(y.east)+(0.4cm, 0.5cm)$)}]
        \node [color=red] (g) at (-0.1,0.1cm) {$g$};
        \pgfplotstableread{
          x y
          0 0.539509442
          1 0.499994381
          2 0.529669216
          3 0.522970023
          4 0.508403436
          5 0.524922819
          6 0.532560819
          7 0.577760101
          8 0.522027899
          9 0.563922241
        }{\loadedtable}
        \begin{axis}[
            ycomb,
            height=5cm,
            width=5cm,
            unit vector ratio*={1 15 0.2},
            clip=false,
            axis lines=none,
            xtick=\empty, ytick=\empty, ztick=\empty,
          ]

          \path [thick,-,color=red] (-0.3,0.53) edge (9.6, 0.53);

          \addplot+[mark size=1.5] table [
              x=x,
              y expr={\thisrow{y} >=0.53 ? \thisrow{y} : NaN},
          ] {\loadedtable};

          \addplot+[mark size=1.5] table [
              x=x,
              y expr={\thisrow{y} < 0.53 ? \thisrow{y} : NaN},
          ] {\loadedtable};

        \end{axis}
      \end{scope}

      \begin{scope}[local bounding box=comb22, shift={($(comb2.east)+(1.6cm, -0.40cm)$)}]
        \pgfplotstableread{
          x y
          0 1.
          1 0.
          2 0.
          3 0.
          4 0.
          5 0.
          6 1.
          7 1.
          8 0.
          9 1.
        }{\loadedtable}
        \begin{axis}[
            ycomb,
            height=5cm,
            width=4.5cm,
            unit vector ratio*={2 5 0.2},
            clip=false,
            axis lines=none,
            xtick=\empty, ytick=\empty, ztick=\empty
          ]

          \draw [ultra thin, -] (axis cs:\pgfkeysvalueof{/pgfplots/xmin},0.)
              -- node[left=1.6cm] {0} (axis cs:\pgfkeysvalueof{/pgfplots/xmax},0.);

          \draw [ultra thin, -] (axis cs:\pgfkeysvalueof{/pgfplots/xmin},1.)
              -- node[left=1.6cm] {1} (axis cs:\pgfkeysvalueof{/pgfplots/xmax},1.);


          \addplot+[mark size=1.5] table [
              x=x,
              y expr={\thisrow{y} >0. ? \thisrow{y} : NaN},
          ] {\loadedtable};

          \addplot+[mark size=1.5] table [
              x=x,
              y expr={\thisrow{y} <= 0. ? \thisrow{y} : NaN},
          ] {\loadedtable};

        \end{axis}
      \end{scope}

      \draw [->, thick] (comb2)  to (comb22);
  \end{tikzpicture}
  \caption{This illustrates how the two neurons with indices $i$ and $j$ from a network produce a series of activations. The $g$ threshold is the red line that splits these activations into two groups: the ones above the threshold (blue) and the ones below the threshold (red). They correspond to the $o_{i}^{r,n}$ and $o_{j}^{r,n}$ time series, which produce the time series of binary values $s_{i}^{r,n}$ and $s_{j}^{r,n}$ used to calculate the TE. The process is applied to all pairs of connected neurons.}
  \label{fig:tscompute}
\end{figure}
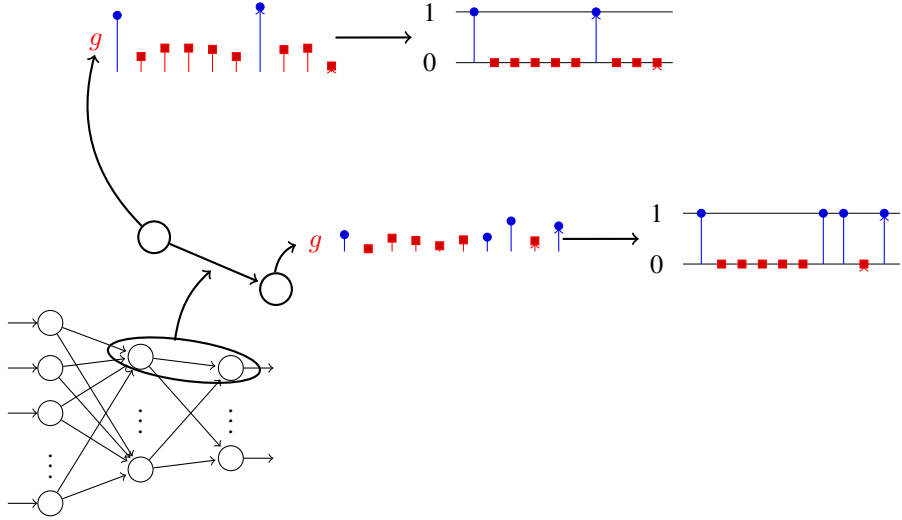

The TE is in general non-negative. However, at certain steps during training, some local TE values can be negative due to noisy inputs resulted from neuron's outputs, and also due to finite training samples available \cite{Prokopenko2013}.

\section{Integrate the TE in the Training of Neural Networks} \label{our_approach}

In the following we explain how we integrate the feedback TE in the training algorithm of a feedforward neural network. This is the key concept of our approach. As a first step, we focus on a one hidden layer perceptron architecture trained with backpropagation gradient descent for classification tasks. Our results can be extended for any number of layers.

Backpropagation \cite{Rumelhart1986} is an algorithm for supervised training of artificial neural networks using gradient descent. It stands for ``backward propagation of errors''. Using the gradient of an error function with respect to the network's weights, we calculate this gradient backwards through the network's layers, from the output layer back to the input layer. This process presents all the training set items to the network's input, while iteratively updating the weights and calculating the $te$ values that we use during the weights updates step as shown in the following sections.

The \emph{FF+FB} training algorithm uses the discretized outputs of the neurons from adjacent layers to construct the time series needed for the TE computation. Once obtained, we use it in the backpropagation weight update process. The weights are updated incrementally (online), after processing each input pattern.

The \emph{FF+FB} requires two training stages. Both stages are using the standard backpropagation algorithm with the modifications we describe below. In Stage I, we train the \emph{FF+FB} network with all training samples while re-evaluating and using TE after each sample. At the end of the Stage I training, we store the $te$ values for all the neuron pairs. In Stage II, we train the same network, using the $te$ values computed in Stage I. In summary:

\begin{enumerate}[label=(\Roman*)]
    \item	Train with the TE feedback. During this stage, we apply the calculated TE values after processing each input pattern.
    \item	Train with the TE feedback using the TE values calculated in Stage I, for $R$ epochs.
\end{enumerate}

The \emph{FF+FB} network requires an additional hyperparameter $g$, the threshold used to bin the output of the neurons when generating the time series that is also obtained using grid search. A unique value of the threshold $g$ is used for all neurons.

The value $te_{j,i}^{r,n}$ is calculated using neurons $j$ and $i$ located in layer $l$, respectively, $l-1$, for the $n^{th}$ sample at epoch $r$.
Using this notation and lag one in the definition of TE (see eq (\ref{eq:TEcond})), we obtain the TE measure of interneuron connections:

\begin{equation}\label{eq:TElagone}
  te_{j,i}^{r,n}=\sum_{s_{i}^{r,n+1}, \; s_i^{r,n}, \; s_j^{r,n}}{p(s_{i}^{r,n+1}, \; s_i^{r,n}, \; s_j^{r,n})} \; log \frac{p(s_{i}^{r,n+1}, \; s_i^{r,n}, \; s_j^{r,n}) \; p(s_i^{r,n})} {p(s_{i}^{r,n+1}, \; s_i^{r,n}) \; p(s_i^{r,n}, \; s_j^{r,n})}
\end{equation}
where $s_{i}^{r,n}$ and $s_{j}^{r,n}$ are the time series obtained from the outputs of neurons $i$ and $j$, located in layers $l-1$, $l$, using the $n^{th}$ sample at epoch $r$.

Once Stage I is completed, we hold only the most recent $te_{j,i}^{r,n}$ values for updating the weights in Stage II. For given values of $r$ and $n$, we simplify the notation by removing indices $r$ and $k$ and introduce the layer index $l$. Instead of $te_{j,i}^{r,n}$, we use $te_{j,i}^{l}$, which aligns with the standard backpropagation notations. We update the weights by a modified gradient descent:
\begin{equation}\label{eq:deltaw}
  \Delta w_{ij}^{l} = -\eta \frac{\partial C}{\partial w_{ij}^{l}} (1 - te_{j,i}^l)
\end{equation}
where $C$ is the loss function. This is one of the changes we have made to the backpropagation algorithm, the addition of $1 - te_{j,i}^l$ term in the calculation of the network's weights updates. There are many possible feedback loop modifications of the backpropagation algorithm, but ours has been experimentally proven to yield positive results compared with the standard algorithm.

We include these modifications in the standard backpropagation gradient descent algorithm \cite{Rumelhart1986}, and obtain Algorithm ~\ref{alg:tebackprop}, used to train the \emph{FF+FB}. This represents the standard backpropagation algorithm for a single hidden layer perceptron network; our additions are at lines \emph{13, 14, 23--26}.

\begin{algorithm}[H]
\newcommand\mycommfont[1]{\footnotesize\ttfamily{#1}}
\SetCommentSty{mycommfont}
\LinesNumbered
\DontPrintSemicolon
\SetKwData{TR}{$g$}
\SetKwComment{Comment}{$//$}{}
\SetKwData{MaxEpochs}{$R$}
\SetKwData{trainingset}{training set}
\SetKwData{netactivations}{$z^{l,n}$}
\SetKwData{Layer}{layer}
\SetKwData{OutputLayer}{output layer}
\SetKwData{HiddenLayer}{hidden layer}
\SetKwData{Sample}{sample}
\SetKwData{Epoch}{epoch}
\SetKwData{SamplesCount}{$N$}
\SetKwData{Neuron}{neuron}
\SetKwData{Weights}{$\pmb{W}$}
\SetKwData{Eta}{\eta}
\SetKwData{TrainStage}{Training Stage}
\SetKwData{OS}{${o}^{l,r,n}_{i,j}$}
\SetKwData{TS}{${s}^{l,r,n}_{i,j}$}
\SetKwData{TE}{${te}^{l}$}
\SetKwInOut{Input}{input}
\SetKwInOut{Output}{output}
\SetKwFunction{Activate}{$w^l\sigma(z^{l-1})+b^l$}
\SetKwFunction{Sigmoid}{$\frac{\mathrm{1}}{\mathrm{1}+e^{-x}}$}
\SetKwFunction{Cost}{$J(\pmb{W})$}
\SetKwFunction{CostDef}{$J(\pmb{W})=-\mathlarger{\sum_{n=1}^{N}{y^{n}\log{(a^{n}) + (1 - y^{n}) \log{(1-a^{n})}}}}$}
\Begin{
  - \Input{$\pmb{y}$ true class labels vector, $\pmb{x}$ input vector, $N$ number of training samples, $R$ number of epochs, $g$ the threshold rate}
  - \Input{\emph{TrainStage} takes the following values: \emph{I}, \emph{II}
  }
  - initialize all \Weights weights with random samples between 0 and 1, drawn from a 0 centered, 0.1 width, normal distribution\;
  - initialize all biases $\pmb{b}$ and activations $\pmb{A}$ with 0.0\;
  - initialize \TS, $i$ and $j$ are the \Neuron indexes of the ${l-1}^{th}$, $l^{th}$ \Layer respectively, and $k$ is the \Sample's index in the \trainingset, $r$ - current epoch \;
  \ForEach{\Epoch $r=0$ \KwTo \MaxEpochs}{
    - randomize \trainingset\;
    \ForEach{\Sample $k$ in \trainingset}{
      \ForEach{$l$ \Layer}{
        - compute input layer outputs $z^{l,k}= \pmb{x} \odot \pmb{W}^{(in)} + b^l$\;
        - compute \netactivations=\Activate for hidden and output layers, where $\sigma(x)=\Sigmoid$\;
        \lIf{\TrainStage \emph{I} {\bf and} $\netactivations < \TR$}{$\TS = 0$}
        \lElse{$\TS = 1$}
        \uIf{$l$ is \HiddenLayer}{
          - compute output error vector $\pmb{\delta}^{(hidden)} = \pmb{\delta}^{(out)} {\pmb{W}^{(out)}}^{T} \odot \cfrac{\partial \sigma{(z^{(hidden)})}} {\partial z^{(hidden)}}$;\
          \Comment*{$\cfrac{\partial \sigma{(z^{(hidden)})}} {\partial z^{(hidden)}} = \sigma{(z^{(hidden)})} \odot (1 - \sigma{(z^{(hidden)})})$ the derivative of the activation function}\;
          - compute derivation of the \Cost function $\cfrac{\partial}{\partial w_{i,j}^{(hidden)}}J(\pmb{W}) = \sigma{(z^{(input)}_{j})} \delta^{(out)}_{i}$, vectorized as:
          $\Delta^{(hidden)} = \Delta^{(hidden)} + (\pmb{A}^{(in)})^T \delta^{(hidden)}$
        }
        \ElseIf{$l$ is \OutputLayer}{
          - compute output error vector $\pmb{\delta}^{(out)} =\netactivations^{(out)} - \mathbf{y}$\;
          - compute derivation of the \Cost function $\cfrac{\partial}{\partial w_{i,j}^{(out)}}J(\pmb{W}) = \sigma{(z^{(hidden)}_{j})} \delta^{(out)}_{i}$, vectorized as:
          $\Delta^{(out)} = \Delta^{(out)} + (\pmb{A}^{(hidden)})^T \delta^{(out)}$
        }
        \uIf{\TrainStage \emph{I} {\bf and} $r=\MaxEpochs$ {\bf and} $k=\SamplesCount$}{
          - compute ${te}_{i,j}^{l}$ using \TS according to \eqref{eq:TElagone}
        }
        \uIf{\TrainStage \emph{I} {\bf or} \TrainStage \emph{II}} {
          $\pmb{W}^{l} := \pmb{W}^{l} - \eta \Delta^{l} (1 -\pmb{{te}}^{l})$
        }
      }
    }
  }
}
\caption{Backpropagation using transfer entropy}
\label{alg:tebackprop}
\end{algorithm}

As mentioned before, the closest related work is Herzog's et al. paper \cite{Herzog2017}. The differences are as follows.

\begin{itemize}[leftmargin=*,labelsep=5.5mm]
  \item In contrast to our method, in \cite{Herzog2017} the computed TE values are used only in their last training step as an input in the activation function; the activation function is $\tilde{g}(x_i)=g(x_i + \sum\nolimits_j{ f_{j \rightarrow i}})$, where $f_{j \rightarrow i}$ is $(w_{min} |\beta - \alpha| )$ divided by the layer count ($\beta$ and $\alpha$ are layer indices, $\beta > \alpha$, and $w_{min}$ is the smallest weight value in their network determined in the first training step). The $f_{j \rightarrow i}$ feedback is only used if the averaged by class TE value is below a threshold $\Phi$.
	\item In our approach, the distance between layers is not required since we consider only pairs of neurons from neighboring layers.
	\item In \cite{Herzog2017}, only larger $te$ values are used in the training stage. We use all computed $te$ values. This adaptation helps us to obtain a longer series of events, which experimentally showed to improve the training process.
\end{itemize}

\section{Experimental Results} \label{results}

The main question is if the addition of the TE factor improves learning and creates any benefits. To answer it, we compare the \emph{FF+FB} and \emph{FF} algorithms using a target accuracy.

The experimental set-up is the following. We define a fixed target validation accuracy that both networks have to reach in an also fixed maximum number of epochs. The learning process ends when either of the target validation accuracy or the maximum number of epochs is reached. The hyperparameters used are the standard ones in any multilayer perceptron (MLP): learning rate $\eta$, number of epochs $R$, and number of neurons in each layer. To make comparisons easier, we use the same learning rate $\eta$ for \emph{FF+FB} and \emph{FF}, determined by grid search. In addition, we use the target accuracy (which is the early training stop limit). For the \emph{FF+FB} we also use the binning threshold hyperparameter $g$.

First, we train the \emph{FF+FB} network on a toy example: the XOR (or ``exclusive or'') problem. The XOR problem is a classic benchmark in neural network research. It is the problem of using a neural network to predict the outputs of XOR logic gates given two (or more) binary inputs. An XOR function should return a true value if the inputs are not equal and a false value if they are equal. We use the most simple XOR problem, with two inputs.

The accuracy measured is here the training accuracy---the one obtained on the training set. Therefore, for now we focus only on the learning cycles, disregarding the overfitting/generalization aspect. We train up to saturation, or to $100\%$ training accuracy the \emph{FF+FB} on the XOR dataset (Table \ref{tab:xordataset}), observing the number of epochs required to reach this accuracy, in comparison to the \emph{FF} network.

\begin{table}[H]
  \centering
  \caption{The XOR dataset. A training epoch consists of 200 vectors, randomly selected from this dataset.}
  \begin{tabular}{ccc}
  \toprule
  \textbf{input 1}	& \textbf{input 2}	& \textbf{output}\\
  \midrule
  0		& 0		  & 0\\
  0		& 1			& 1\\
  1		& 0		  & 1\\
  1		& 1			& 0\\
  \bottomrule
  \end{tabular}
  \label{tab:xordataset}
\end{table}

The number of epochs for training the \emph{FF+FB} network to reach the target accuracy is 7--10 times less than for \emph{FF}. We average $10$ runs, taking $g = 0.7$ and $\eta = 0.025$ for both networks. The number of epochs is capped to $300$ epochs. With these constraints, $\eta$ is optimized (by grid search) for the smallest number of epochs of the two networks that reached 100\% training accuracy in 10 runs. The network for the XOR dataset has $2-2-1$ (input--hidden--output) nodes, as depicted in Figure \ref{fig:xorarch}.

\def\layersep{2.5cm}
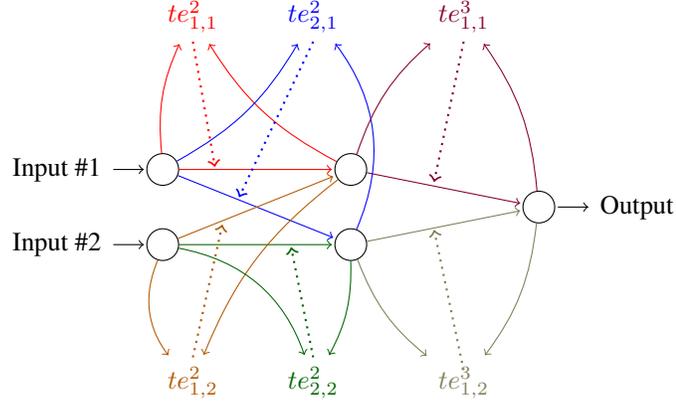
\begin{figure}[H]
  \centering
  \hspace*{50pt}
  \begin{tikzpicture}[shorten >=1pt,->,draw=black, node distance=\layersep]
    \begin{scope}
      \tikzstyle{every pin edge}=[<-,shorten <=1pt]
      \tikzstyle{neuron}=[circle,draw,fill=none,minimum size=12pt,inner sep=0pt]
      \tikzstyle{ann} = [draw=none,fill=none,right]
      \tikzstyle{input neuron}=[neuron];
      \tikzstyle{output neuron}=[neuron];
      \tikzstyle{hidden neuron}=[neuron];
      \tikzstyle{annot} = [text width=4em, text centered]
      \tikzstyle{ann} = [text width=2em, text centered]

      \foreach \name / \y in {1,...,2}
          \node[input neuron, pin=left:Input \#\y] (I-\name) at (0,-\y) {};

      \foreach \name / \y in {1,...,2}
          \path[yshift=0.0cm]
              node[hidden neuron] (H-\name) at (\layersep,-\y cm) {};

      \node[output neuron,yshift=0.5cm, pin={[pin edge={->}]right:Output}, right of=H-2] (O) {};

      \path [red, name path=W1-1] (I-1) edge (H-1);
      \path [blue, name path=W1-2] (I-1) edge (H-2);
      \path [black!30!orange, name path=W2-1] (I-2) edge (H-1);
      \path [black!60!green, name path=W2-2] (I-2) edge (H-2);
      \path [black!30!purple, name path=O1-1] (H-1) edge (O);
      \path [black!60!yellow, name path=O1-2] (H-2) edge (O);

      \node[ann, red] (te112) [above=of H-1,xshift=-0.8cm] at (1.2, -1.8cm) {$te_{1,1}^2$};
      \path[->, red, bend left=15] (H-1) edge (te112);
      \path[->, red, bend left=15] (I-1) edge (te112);
      \draw [->, red, bend right=10, dotted, thick]
        ($ (te112.south) + (0mm, -0.) $)
        -- ++ (0.3, -1.7);

      \node[ann, blue] (te212) [above=of H-2,xshift=+0.8cm] at (1.2, -1.8cm) {$te_{2,1}^2$};
      \path[->, blue, bend right=30] (H-2) edge (te212);
      \path[->, blue, bend right=15] (I-1) edge (te212);
      \draw [->, blue, bend right=25, dotted, thick]
        ($ (te212.south) + (0mm, -0.) $)
        -- ++ (-1.0, -2.1);

      \node[ann, black!30!orange] (te122) [below=of H-2,xshift=-0.8cm] at (1.2, -1.0cm) {$te_{1,2}^2$};
      \path[->, black!30!orange, bend right=15] (H-1) edge (te122);
      \path[->, black!30!orange, bend right=30] (I-2) edge (te122);
      \draw[->, black!30!orange, bend right=25, dotted, thick]
        ($ (te122.north) + (0mm, -0.) $)
        -- ++ (+0.4, +1.8);

      \node[ann, black!60!green] (te222) [below=of H-2,xshift=+0.8cm] at (1.2, -1.0cm) {$te_{2,2}^2$};
      \path[->, black!60!green, bend left=15] (H-2) edge (te222);
      \path[->, black!60!green, bend left=30] (I-2) edge (te222);
      \draw [->, black!60!green, dotted, thick]
        ($ (te222.north) + (0mm, -0.) $)
        -- ++ (-0.3, +1.5);

      \node[ann, black!30!purple] (te113) [above=of O,xshift=+2.8cm] at (1.2, -1.8cm) {$te_{1,1}^3$};
      \path[->, black!30!purple, bend left=15] (H-1) edge (te113);
      \path[->, black!30!purple, bend right=15] (O) edge (te113);
      \draw [->, black!30!purple, bend right=25, dotted, thick]
        ($ (te113.south) + (0mm, -0.) $)
        -- ++ (-0.4, -1.9);

      \node[ann, black!60!yellow] (te123) [below=of O,xshift=+2.8cm] at (1.2, -1.0cm) {$te_{1,2}^3$};
      \path[->, black!60!yellow, bend right=15] (H-2) edge (te123);
      \path[->, black!60!yellow, bend left=15] (O) edge (te123);
      \draw [->, black!60!yellow, bend right=25, dotted, thick]
        ($ (te123.south) + (0mm, +0.5) $)
        -- ++ (-0.4, +1.9);
    \end{scope}
  \end{tikzpicture}
  \caption{The \emph{FF+FB} architecture for the XOR problem. The $te_{j,i}^l$ values are calculated between neurons $j$ and $i$. Neuron $i$ is in layer $l-1$, neuron $j$ is in layer $l$. The colored arrows from the neurons show how the outputs from the neurons are used to calculate the $te$ values. The same color in a dotted line arrow shows to which weight the $te$ is applied (see Equation \ref{eq:deltaw}). The bias units are implemented but not shown here since they do not use the $te$ values in the algorithm.}
  \label{fig:xorarch}
\end{figure}

\begin{figure}[H]
\centering
\includegraphics[width=36pc]{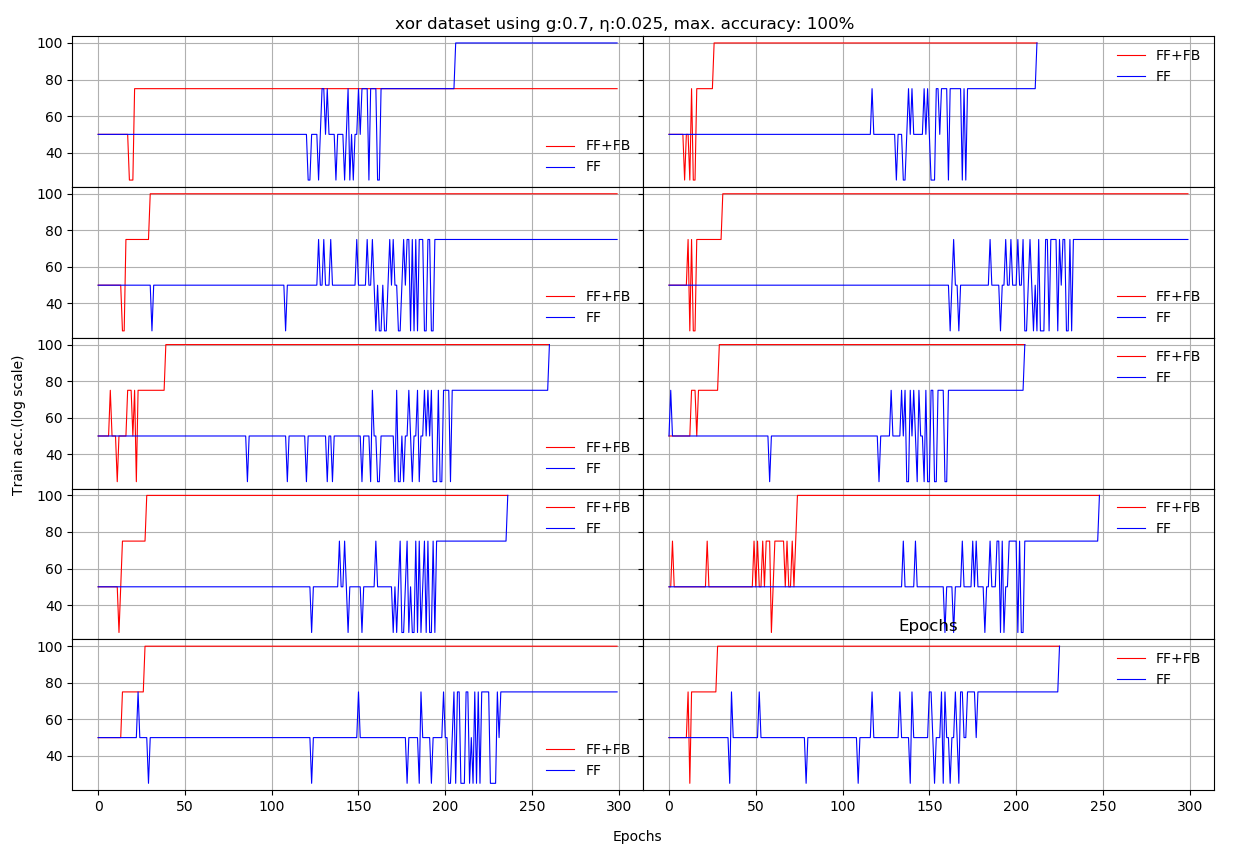}
\caption{Ten runs on XOR dataset. Each \emph{x} axis finishes when the last of the \emph{FF+FB} or \emph{FF} reaches either the maximum number of epochs or 100\% training accuracy (log scale).}
\label{fig:xorruns}
\end{figure}

For the XOR problem, the \emph{FF+FB} network is more efficient for a relatively small $\eta$ (between 0.022 and 0.045) and $g = 0.7$. However, these $\eta$ values are not also optimal for the \emph{FF} network. For a larger learning constant ($\eta \approx 0.1$), \emph{FF} generally needs less epochs (i.e., converges faster). For \emph{FF+FB}, a small $\eta$ ($\eta < 0.45$) paired with a small $g$ ($g < 0.25$) or a large $g$ ($g > 0.7$) threshold destabilizes \emph{FF+FB} training. Figure \ref{fig:xorruns} and Table \ref{tab:xorruns} depicts the evolution of XOR learning for each run.

\begin{table}[H] 
  \caption{Comparison between the number of epochs required by \emph{FF+FB} and \emph{FF} to reach 100\% training accuracy on the XOR dataset in 10 runs. The networks did not successfully reach the target accuracy on runs 1 - for \emph{FF+FB} and runs 2,5,7 for \emph{FF}, showing the maximum number of epochs (300) the training was limited to.}
  \centering

  \begin{tabular}{ccc}
  \toprule
  \textbf{Run}	& \textbf{\emph{FF+FB} epochs at 100\% acc.}	& \textbf{\emph{FF} network epochs at 100\% acc.}\\
  \midrule
  1		    & 300		  & 207\\
  2		    & 32			& 300\\
  3		    & 40  		& 261\\
  4		    & 75			& 249\\
  5		    & 28			& 300\\
  6		    & 27		  & 213\\
  7		    & 31		  & 300\\
  8		    & 30			& 206\\
  9		    & 29			& 237\\
  10	    & 29		  & 226\\
  \textbf{Average} & \textbf{62.2}    & \textbf{349.9}\\
  \bottomrule
  \end{tabular}

  \label{tab:xorruns}
\end{table}

Next, we train the two networks on ten standard datasets \cite{UCI2019}: \emph{abalone}, \emph{car}, \emph{chess}, \emph{glass}, \emph{ionosphere}, \emph{iris}, \emph{liver}, \emph{redwine}, \emph{seeds}, and \emph{divorce}. For these set of experiments, the accuracy is measured on independent test sets, as usual. Each of the datasets had specific target accuracies and number of epochs.

The question is which of the two networks reaches the target accuracy in less epochs. We obtain on almost all datasets increased accuracies in a 10 run average, and for most of them, \emph{FF+FB} reaches the target accuracies in less epochs than the \emph{FF} network.

We noticed that using an unoptimized hidden layer size negatively impacts \emph{FF} network's training. The \emph{FF+FB} is less sensitive to this aspect (up to certain thresholds), and can successfully converge to the target accuracy; however, for some datasets, it requires more epochs.

In the Figures \ref{fig:appafig1}, \ref{fig:appafig2}, \ref{fig:appafig3} from Appendix \ref{trainings} we depicted the evolution of the learning process for all 10 datasets per each run. Our proposed solution (the \emph{FF+FB} model) is generally more stable and reaches the accuracy target in less epochs than the \emph{FF} network.

We also performed several control experiments to assess the significance of these results. We verified if variations of our modified backpropagation, including detrimental and misuse of the $te$, could produce different or similar results. In Section \ref{discussion} we discuss some of these results. We explored the following.

\begin{itemize}[leftmargin=*,labelsep=4.9mm]
      \item	Set a fixed $te$ value for all feedbacks.
      \item	Strengthen/weaken the $te_{j,i}^l$ values by layer index.
      \item Replace all weights with fixed values and use $te$ as feedback.
      \item Scale the $te_{j,i}^l$ values to $[0, 1]$.
\end{itemize}

\section{Discussion} \label{discussion}

Compared to \emph{FF}, the \emph{FF+FB} training algorithm has an overhead needed to compute the $te$ values in Stage I. In Stage II, these values are only used and there is no additional overhead.

According to our experiments, adding the TE feedback parameter (Stage II in \emph{FF+FB}) brings two benefits: \emph{a)} it accelerates the training process---in general, less epochs are needed, and \emph{b)} we generally achieve a better test set accuracy. For the plots shown in Appendix \ref{trainings}, Table \ref{tab:accuracies} summarizes the obtained average accuracies and the differences between \emph{FF+FB} and \emph{FF}.

\begin{table}[H]
  \caption{Comparison of various between \emph{FF+FB} and \emph{FF} for the target validation accuracies on specified datasets (average of 10 runs). Whenever the networks did not successfully reach the targets, we used the maximum number of epochs and the last recorded accuracy to calculate the averages.}
  \centering
  \begin{tabular}{p{3cm} p{1cm} p{1cm} p{1cm} p{1cm} p{1cm} p{1cm} p{1cm}}
  \toprule
  \textbf{Dataset} & \textbf{Target Accuracy} & \textbf{Avg. \emph{FF+FB} Accuracy} & \textbf{\emph{FF+FB} Avg. Epochs} & \textbf{Avg. \emph{FF} Accuracy} & \textbf{\emph{FF} Avg. Epochs} & \textbf{Accuracy Difference} & \textbf{Max Epochs}\\
  \midrule
  abalone     & 52\%  & 53.01	& 6.2 & 52.16 & 37.5 & 0.84\% & 50\\
  car         & 73\%  & 72.21	& 163.5 & 73.14 & 184.2 & -0.92\% & 300\\
  chess       & 96\%  & 96.20 & 19.0 & 95.41 & 38.3 & 0.79\% & 40\\
  glass       & 52\%  & 52.46	& 154.6 & 35.84 & 294.4 & 16.61\% & 300\\
  ionosphere  & 92\%  & 92.17	& 16.4 & 92.26 & 22.5 & -0.09\% & 60\\
  iris        & 92\%  & 95.11	& 13.8 & 96.22 & 24.8 & -1.11\% & 100\\
  liver       & 70\%  & 68.46 & 212.1 & 61.82 & 294.2 & 6.63\% & 300\\
  redwine     & 52\%  & 50.18	& 134.6 & 49.89 & 171.8 & 0.29\% & 200\\
  seeds       & 85\%  & 87.46	& 41.3 & 87.14 & 136.2 & 0.31\% & 200\\
  divorce     & 98\%  & 98.03 & 6.9 & 98.62 & 7.4 & -0.58\% & 20\\
  \bottomrule
  \end{tabular}

  \label{tab:accuracies}
\end{table}

We observed that the optimal $\eta$ values for the \emph{FF+FB} and \emph{FF} networks are different. The \emph{FF+FB} network usually needs a slightly smaller learning rate. As we empirically observed that a smaller learning rate value is more beneficial for the \emph{FF+FB}, we can conclude that \emph{FF+FB} learns in smaller steps. The direction of these steps is in general more targeted, given the smoothness of the accuracy curve for most datasets (see Table \ref{tab:accuracies} and Appendix \ref{trainings}).

Small $\eta$ and $g$ values can make the \emph{FF+FB} network get stuck in local minima. For an unoptimized $\eta$, the \emph{FF+FB} network uses the $te_{j,i}^l$ values to compensate for a poor $\eta$ choice. A good choice of the $g$ threshold becomes more important in this case. Threshold $g$ was determined for each dataset using grid search, after $\eta$ was selected. As we use the sigmoid activation function (with values less than 1), a $g=0.9$ value would mean that only significant activations will be used in the TE computation. The learning rate $\eta$ was selected for each dataset targeting the best results in 10 runs of the \emph{FF} network and the same $\eta$ has been used for \emph{FF+FB}.

Constructing the time series, per epoch, for each training sample independently, does not produce good results: the obtained $te$ values (scaled or not) were very small. Weighting the $te$ values in this scenario, was also not a good approach.

The $te$ values are tuned during Stage I. Calculating $te$ after all training samples were processed proved to be a bad alternative.

Examining the $te$ values raised new questions and we performed additional experiments to alleviate any possible bias in the results. The investigation was motivated by the negative and large $te$ values (observed in most datasets) and their association with increased scores. Using these values in Equation \eqref{eq:deltaw} is not consistent with Equation \eqref{eq:TElagone}, as negative values would mean that the source misinforms the target's next state. By subtracting $te$ from $1$, we revert the negative $te$ values. Since the $te$ values rarely exceed values like +/- 6, this operation is also favorable for the positive extreme values.

For the \emph{car} and \emph{glass} datasets, where learning was slow or capped to inappropriate margins, we used the Weka package (version 3.9.3) with the MultiLayerPerceptron function, with identical or different hyperparameters, as needed, and, where required, with a lot more iterations, to validate our implementation's behavior for an established maximum accuracy. Our implementations of \emph{FF+FB} and \emph{FF} performed at least as good as the ones in Weka.

Computing $te_{j,i}^l$ for all neuron pairs is prohibitive even for shallow networks, especially for training sets with more than $10^5$ samples. However, these are related only to TE computation and occur only during training Stage I. In practice, the trained weights of a neural network can be stored ($te$ values are embedded in these weights). Therefore, in real-world applications, any inference tasks would not be affected by the increased training computational cost for \emph{FF+FB}. Additionally, having the $te$ values obtained in training Stage I stored, they can be reused as needed in training Stage II without any computational overhead. Alternative TE estimation techniques may be considered for such cases \cite{Cataron2019}.

\section{Conclusions} \label{conclusions}

We introduced \emph{FF+FB}, a neural training algorithm which uses information transfer to quantify relationships between neurons and uses the TE feedback to enhance certain neural connections. Our method generally uses less training epochs and achieves higher accuracy compared to the \emph{FF} network. In addition, it is more stable during training, as it can be observed from the plots in Appendix \ref{trainings}, and less sensitive to local minima.

Using the TE feedback can reduce the effort for optimizing hyperparameters like $\eta$ and number of hidden neurons in the hidden layer. According to our experiments, choosing an optimized value for the threshold parameter $g$ can decrease the importance of other hyperparameters (e.g., $\eta$ and the number of hidden nodes).

Our approach could facilitate the extraction of knowledge (and explanations) from the trained networks using the causality paradigm. This is left as an open problem.

\section{Abbreviations}
\label{abbreviations}
The following abbreviations are used in this manuscript:\\

\noindent
\begin{tabular}{p{1 cm} p{12 cm}}
$te$ & transfer entropy value\\
$g$ & the binning threshold\\
$te_{j,i}^{l}$ & obtained $te$ calculated using the time series produced by binning the outputs of neurons' index $i$ from layer $l-1$ and index $j$ from layer $l$\\
$te_{j,i}^{r,n}$ & obtained $te$ using the time series from outputs of neurons' index $i$ from layer $l-1$ and index $j$ from layer $l$, at epoch $r$ and sample $n$\\
$o_i^{r,n}$ & output of neuron having index $i$ - that is layer $l-1$, at epoch $r$ using sample $n$\\
$o_j^{r,n}$ & output of neuron having index $j$ - that is layer $l$, at epoch $r$ using sample $n$\\
$s_i^{r,n}$ & binning the output of neuron index $i$ - layer $l-1$, at epoch $r$ using sample $n$, with threshold $g$\\
$s_j^{r,n}$ & binning the output of neuron index $j$ - layer $l$, at epoch $r$ using sample $n$, with threshold $g$\\
CNN & Convolutional Neural Network\\
\emph{FF+FB} & Feedback Transfer Entropy - our proposed method\\
MLP & Multi-layer perceptron\\
\emph{FF} & Non Feedback network, regular MLP architecture and algorithm\\
TE & Transfer Entropy\\
\end{tabular}

\section{Acknowledgements}
This article was first published in MDPI Entropy Journal on \DTMdisplaydate{2020}{01}{16}{-1}. A.M, A.C. and R.A. equally contributed to the published work. The costs to publish in open access were covered by Siemens SRL. The authors declare no conflict of interest.


\bibliographystyle{unsrt}
\bibliography{references}  



\appendix

\section{Validation Accuracy Evolution during Stage II} \label{trainings}

We present here the results of 10 runs on 10 standard benchmarks by illustrating the evolution of Stage II. We compare the dynamic behavior of \emph{FF+FB}(in red) and \emph{FF}(in blue) for a given target accuracy, with respect to the number of epochs required to reach that target. For most datasets, \emph{FF+FB} improves its accuracy earlier and with a faster rate than \emph{FF}.

\begin{figure}[H]
  \centering
  \includegraphics[width=36pc]{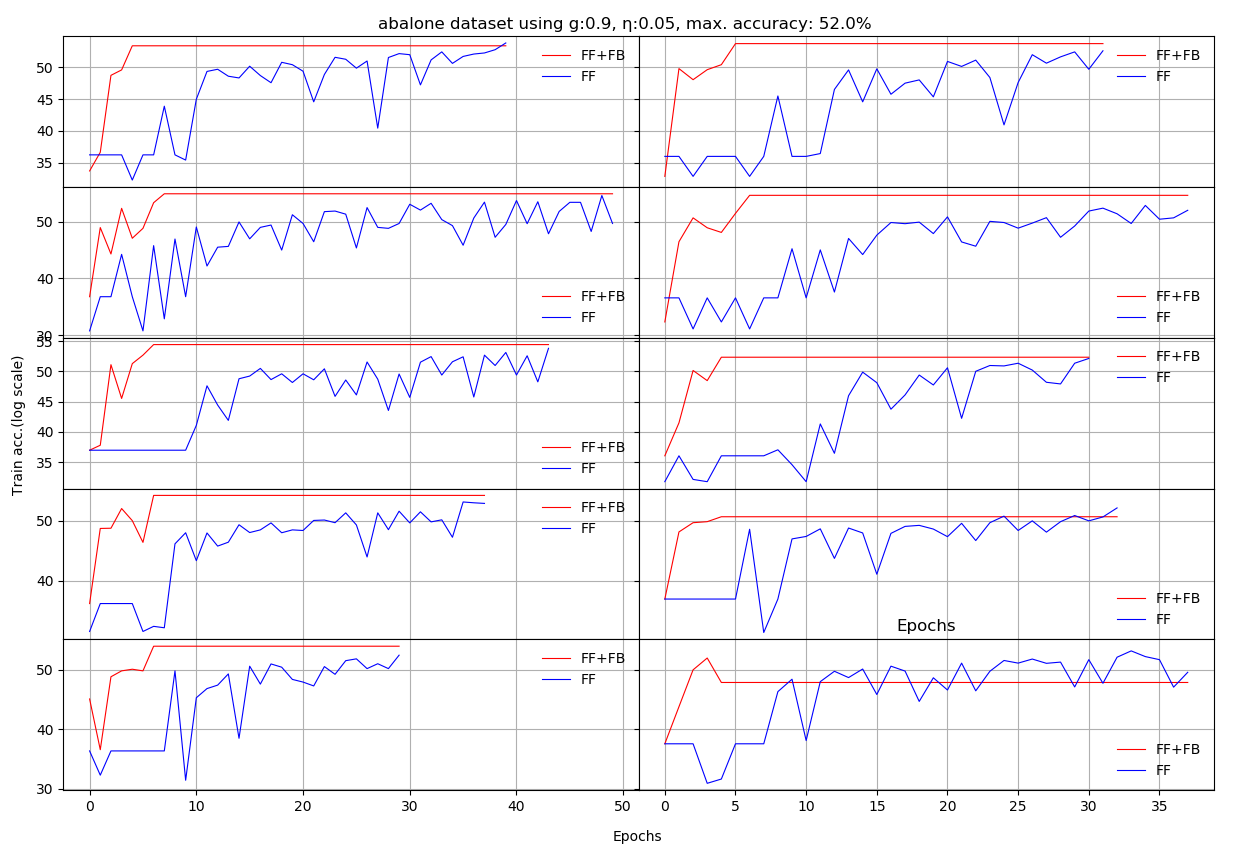}
  \caption{Ten runs of Abalone dataset. Each X axis finishes when the last of the \emph{FF+FB} or \emph{FF} reaches either the maximum number of epochs or the maximum set validation accuracy (log scale).}
  \label{fig:appafig1}
\end{figure}

\begin{figure}[H]
\centering
\includegraphics[width=36pc]{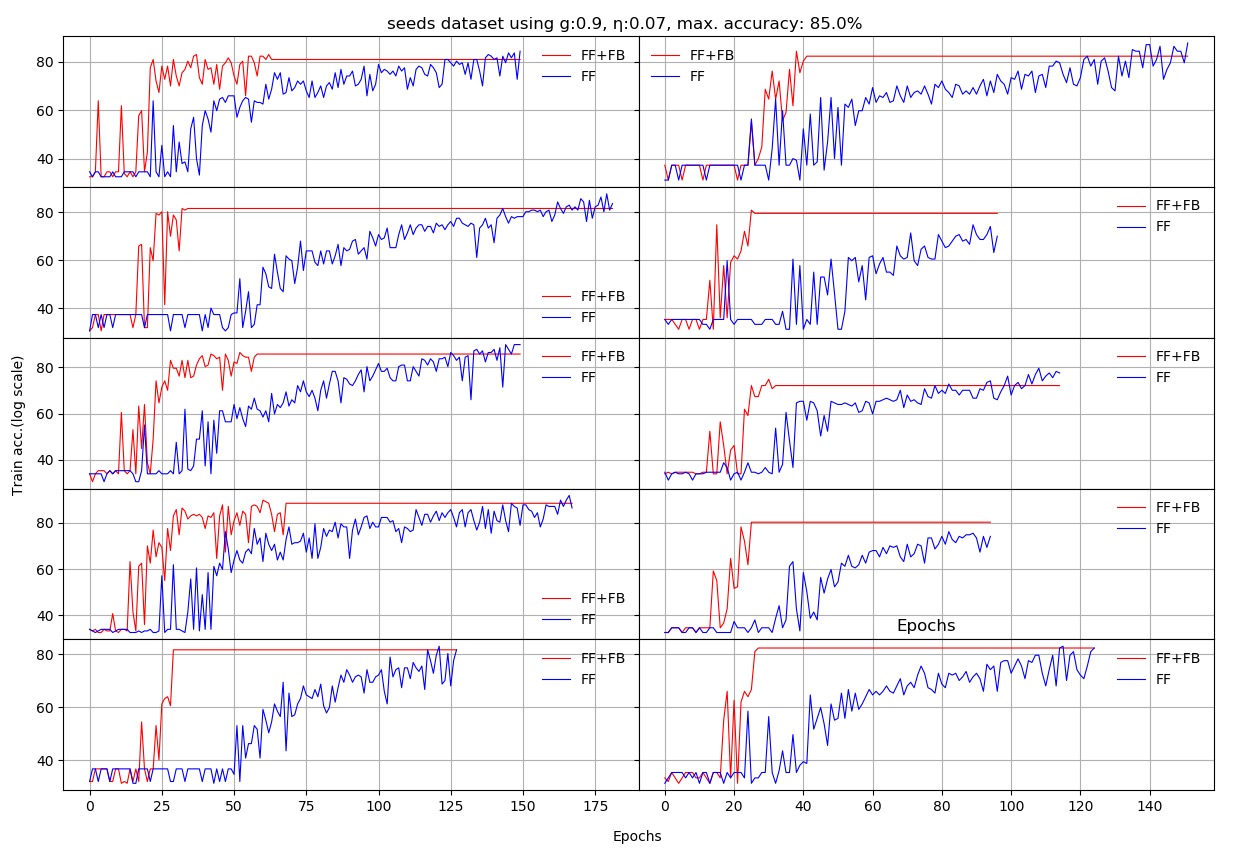}
\end{figure}

 \begin{figure}[H]
  \centering
  \includegraphics[width=36pc]{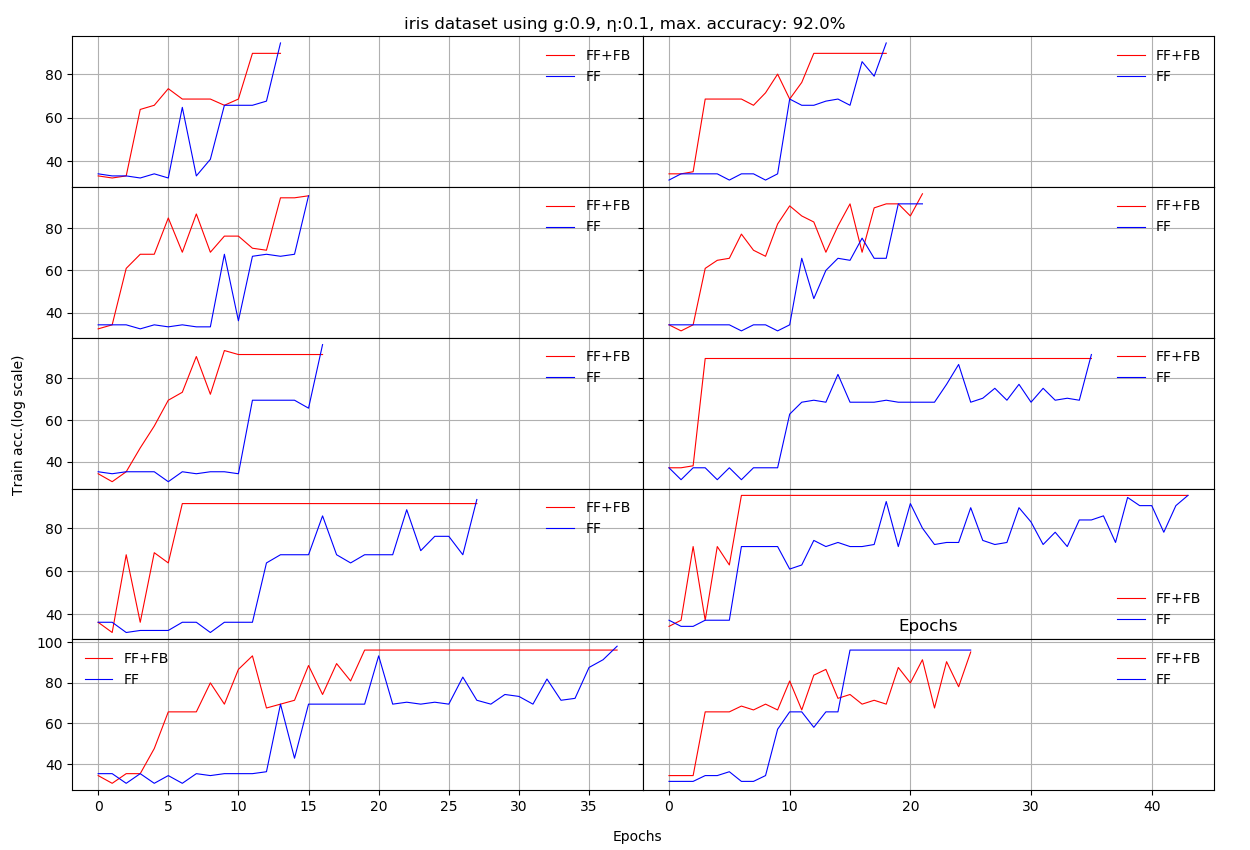}
 \end{figure}

 \begin{figure}[H]
  \centering
  \includegraphics[width=36pc]{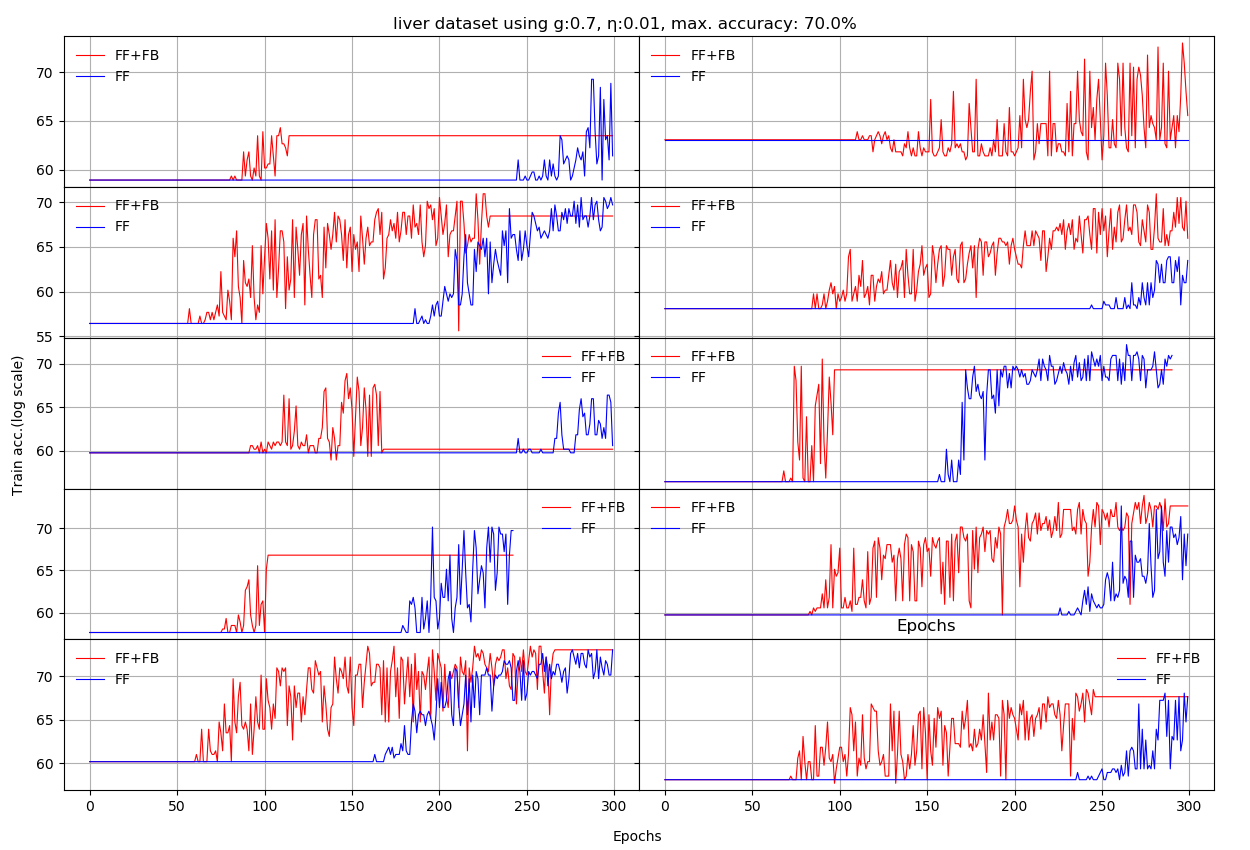}
 \end{figure}

 \begin{figure}[H]
  \centering
  \includegraphics[width=36pc]{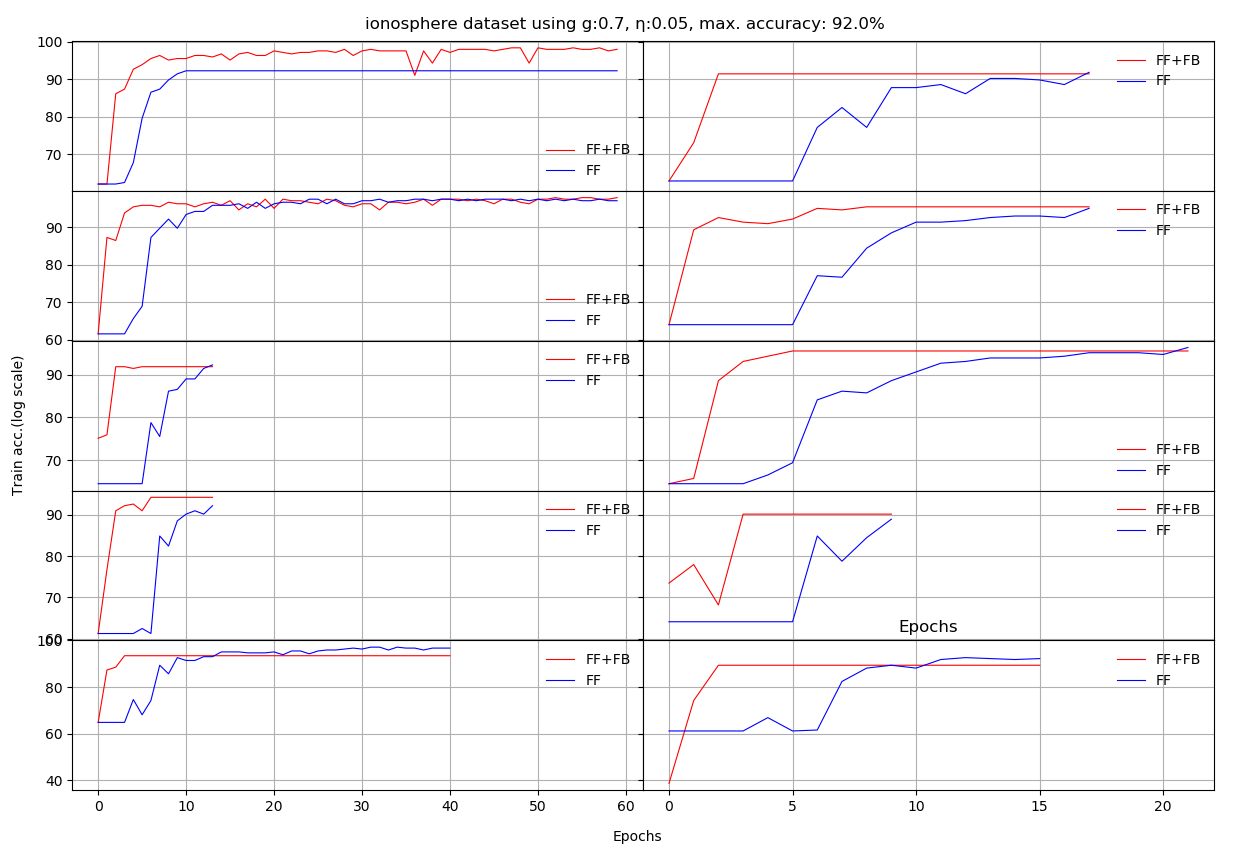}
 \end{figure}

 \begin{figure}[H]
  \centering
  \includegraphics[width=36pc]{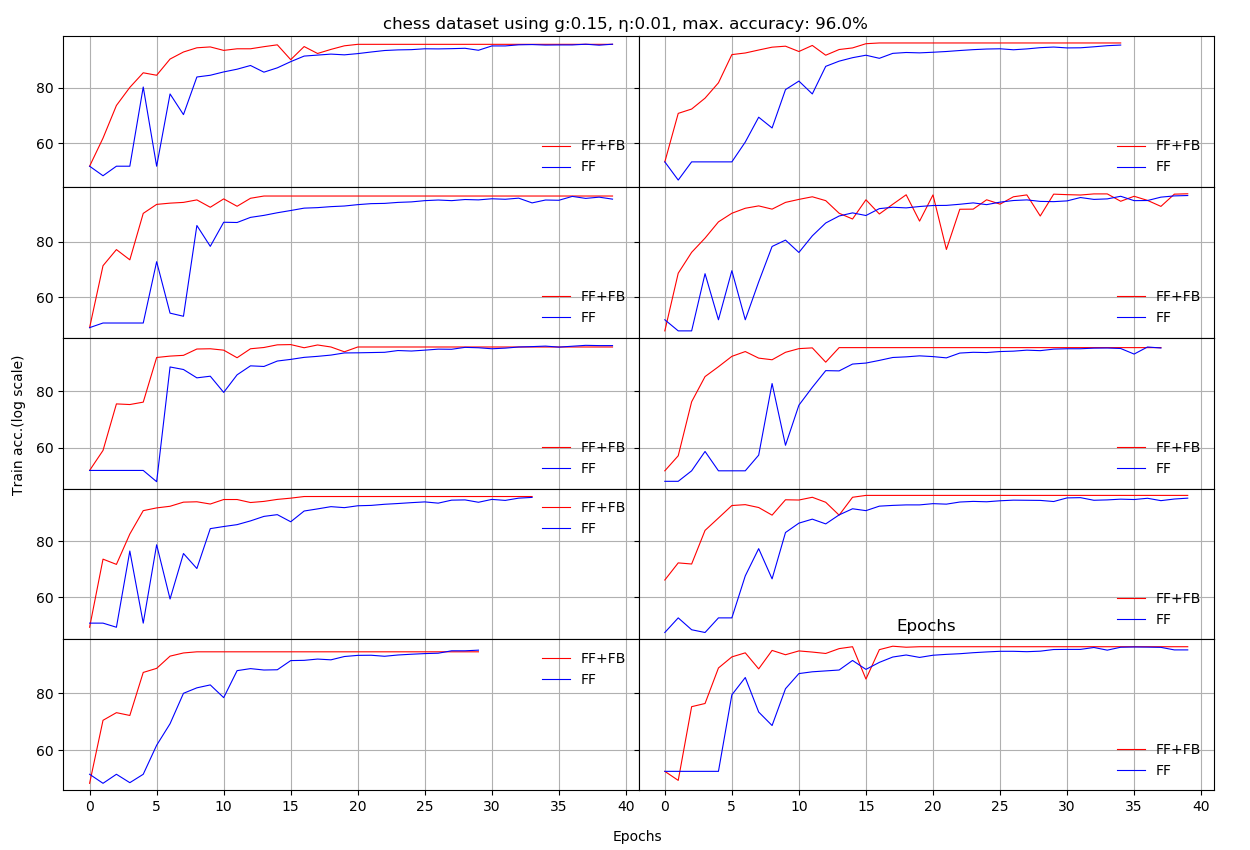}
 \end{figure}

 \begin{figure}[H]
  \centering
  \includegraphics[width=36pc]{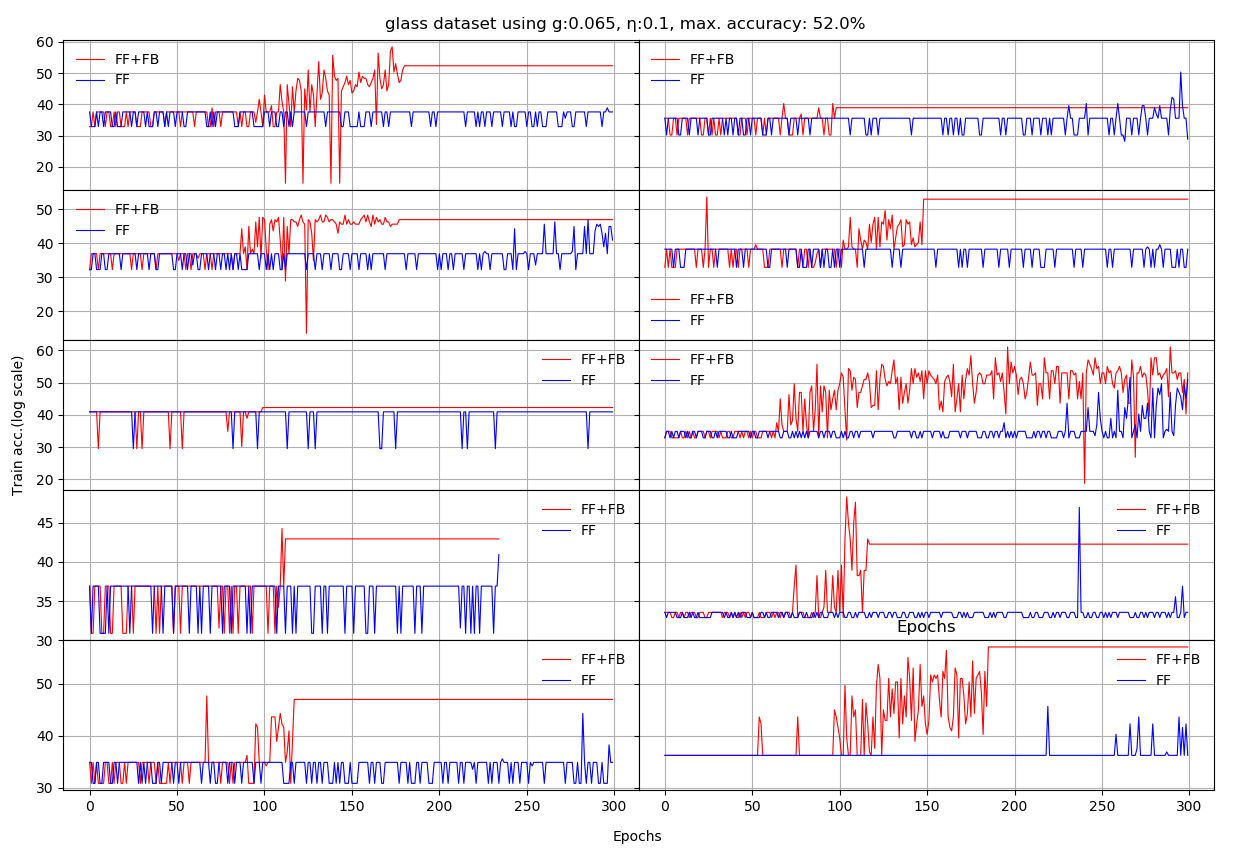}
 \end{figure}

 \begin{figure}[H]
  \centering
  \includegraphics[width=36pc]{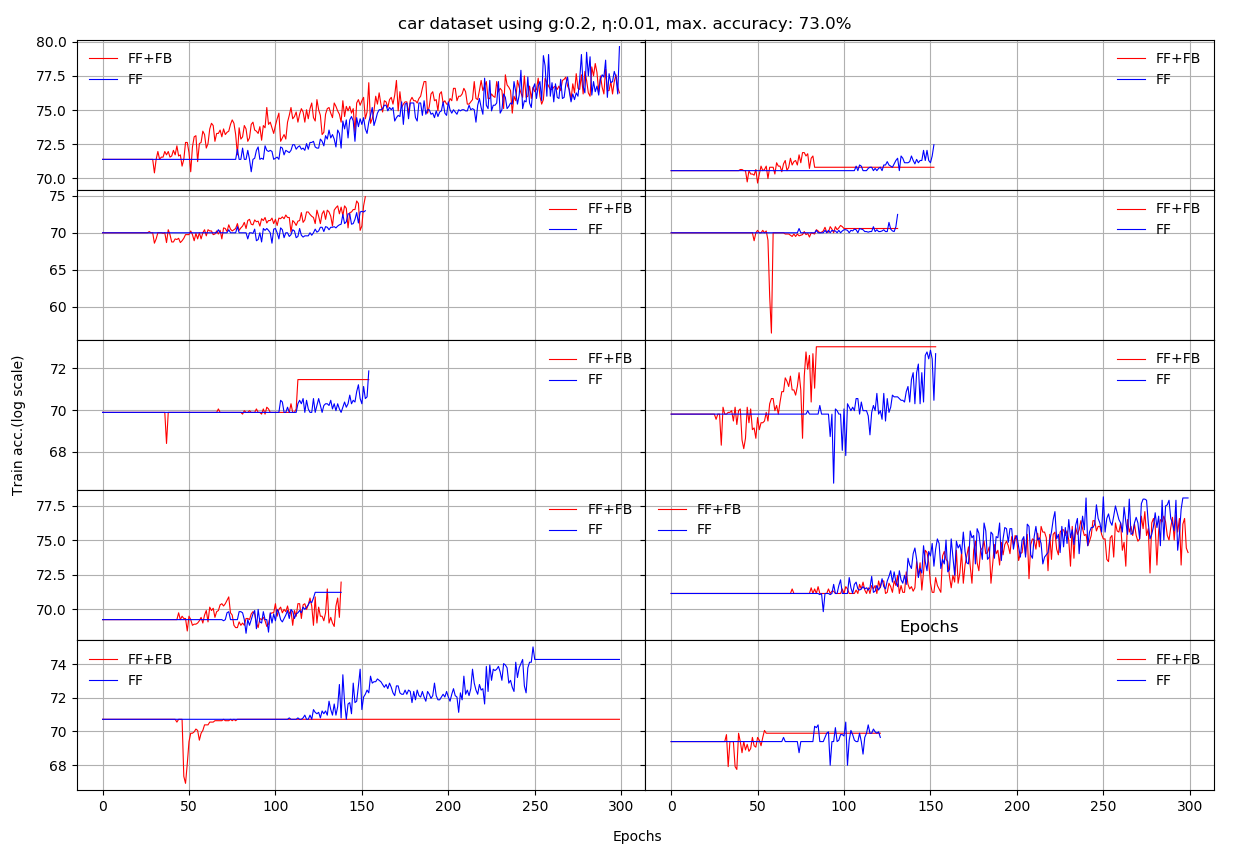}
  \caption{We tried several $\eta$ values for this benchmark. To improve stability of both models, we selected a smaller value. It can be observed that \emph{FF+FB} failed to converge on the 9th run and even to properly learn on other runs.}
  \label{fig:appafig2}
 \end{figure}

 \begin{figure}[H]
  \centering
  \includegraphics[width=36pc]{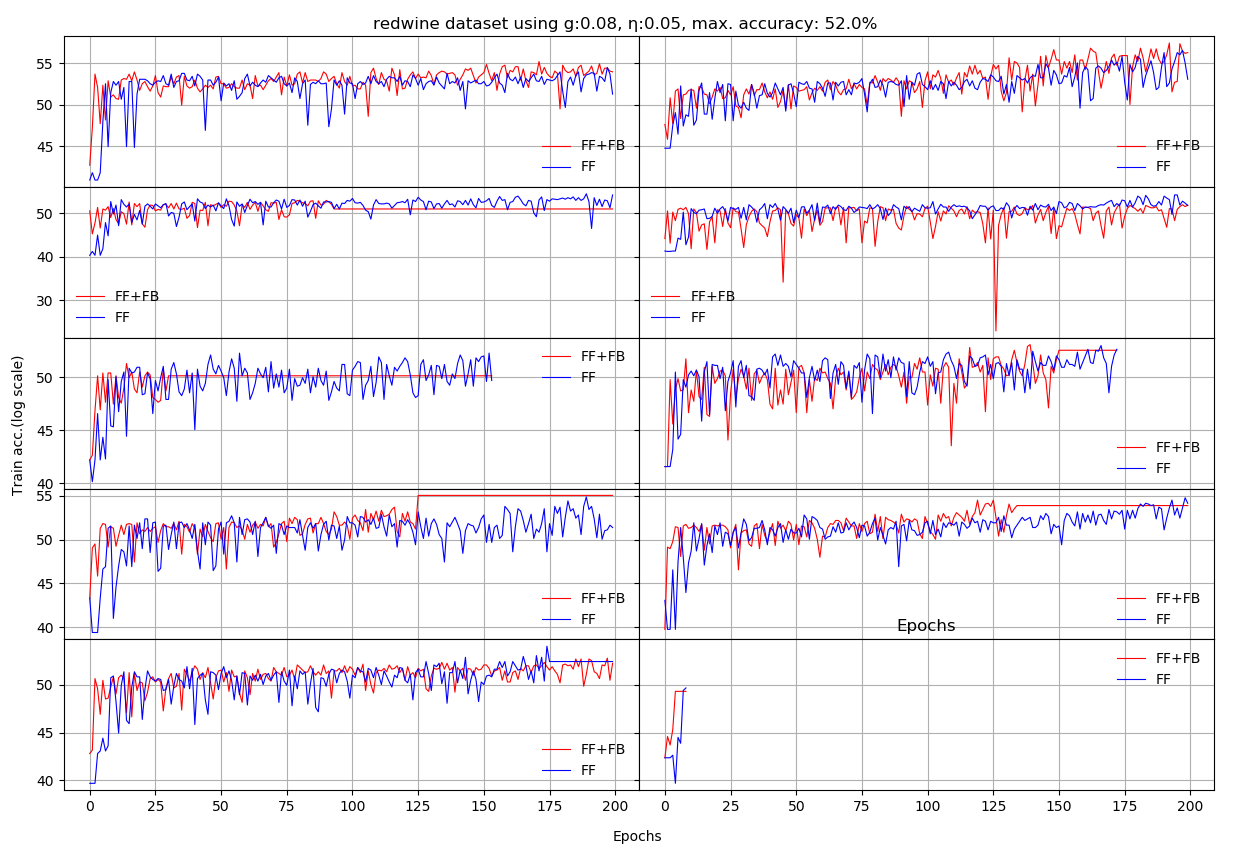}
 \end{figure}

 \begin{figure}[H]
  \centering
  \includegraphics[width=36pc]{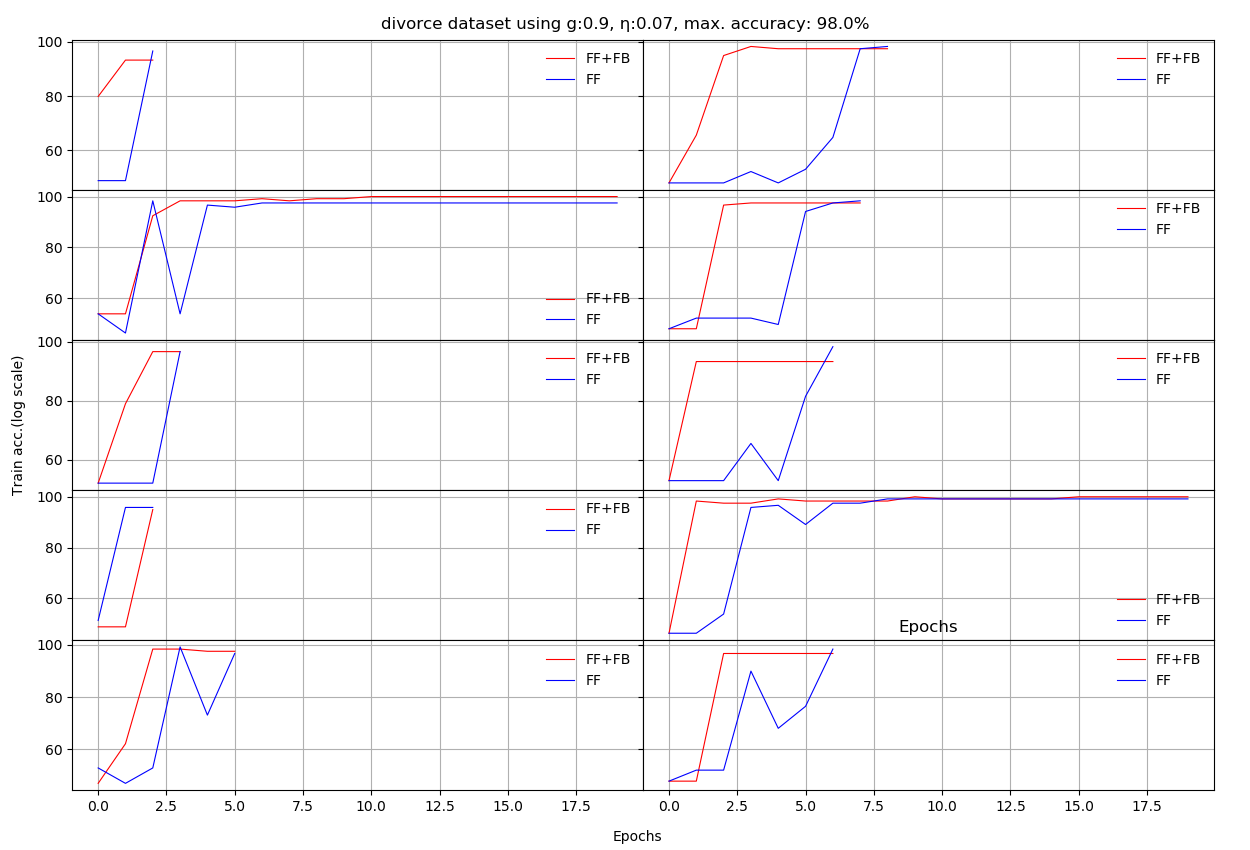}
  \caption{The \emph{divorce} dataset constantly requires only a few epochs to reach the target accuracy.}
  \label{fig:appafig3}
 \end{figure}

\end{document}